\documentclass[runningheads]{llncs}
\usepackage[T1]{fontenc}
\usepackage{graphicx}
\usepackage{booktabs}
\usepackage[misc]{ifsym}
\usepackage{amsmath} 
\usepackage{tcolorbox}
\usepackage{adjustbox}
\usepackage{multirow}
\usepackage{float}
\usepackage{amsfonts}
\usepackage[hidelinks]{hyperref} 

\usepackage{mwe}
\begin{document}
\newcommand{\method}{{LLM-GOOD}\xspace}

\title{Few-Shot Graph Out-of-Distribution Detection with LLMs}

\titlerunning{Few-Shot Graph Out-of-Distribution Detection with LLMs}


\author{Haoyan Xu\inst{1}\textsuperscript{*}  \and
Zhengtao Yao \inst{1}\textsuperscript{*}  \and
Yushun Dong\inst{2} \and
Ziyi Wang\inst{2} \and
Ryan A. Rossi\inst{2} \and
Mengyuan Li\inst{1} \and
Yue Zhao \inst{1}}

\authorrunning{H. Xu et al.}

\institute{University of Southern California, \email{\{haoyanxu, zyao9248, mengyuanli, yzhao010\}@usc.edu}
\and
Florida State University \email{yushun.dong@fsu.edu}
\and
University of Maryland, College Park
\email{zoewang@umd.edu}
\and
Adobe Research
\email{ryrossi@adobe.com}}

\maketitle              
\renewcommand{\thefootnote}{\fnsymbol{footnote}}
\footnotetext[1]{Equal contribution.}
\begin{abstract}
Existing methods for graph out-of-distribution (OOD) detection typically depend on training graph neural network (GNN) classifiers using a substantial amount of labeled in-distribution (ID) data. However, acquiring high-quality labeled nodes in text-attributed graphs (TAGs) is challenging and costly due to their complex textual and structural characteristics. Large language models (LLMs), known for their powerful zero-shot capabilities in textual tasks, show promise but struggle to naturally capture the critical structural information inherent to TAGs, limiting their direct effectiveness.

To address these challenges, we propose \method, a general framework that effectively combines the strengths of LLMs and GNNs to enhance data efficiency in graph OOD detection. Specifically, we first leverage LLMs' strong zero-shot capabilities to filter out likely OOD nodes, significantly reducing the human annotation burden. To minimize the usage and cost of the LLM, we employ it only to annotate a small subset of unlabeled nodes. We then train a lightweight GNN filter using these noisy labels, enabling efficient predictions of ID status for all other unlabeled nodes by leveraging both textual and structural information. After obtaining node embeddings from the GNN filter, we can apply informativeness-based methods to select the most valuable nodes for precise human annotation. Finally, we train the target ID classifier using these accurately annotated ID nodes.
Extensive experiments on four real-world TAG datasets demonstrate that \method significantly reduces human annotation costs and outperforms state-of-the-art baselines in terms of both ID classification accuracy and OOD detection performance.

\end{abstract}

\section{Introduction}
Out-of-distribution (OOD) detection \cite{liu2020energy,hendrycks2016baseline} has emerged as a critical task in machine learning, particularly for safety-critical applications where models must reliably identify inputs that differ significantly from the training data. Recently, several OOD detection methods \cite{song2022learning,wu2023energy} and open-set learning approaches \cite{wu2020openwgl,xu2024lego} have been proposed and applied to graph-structured data. 
Existing graph OOD detection methods typically operate within a semi-supervised, transductive framework, where the entire set of nodes is accessible during training, but only a portion of the class labels (in-distribution (ID) classes) are provided \cite{song2022learning}. These methods generally rely on a sufficient number of labeled ID nodes to train a GNN-based ID classifier, from which they derive the ID classification logits for all nodes. Post-hoc OOD detectors \cite{wu2023energy,hendrycks2016baseline,marevisiting} are then applied to these logits for OOD detection. In particular, nodes with higher energy scores \cite{wu2023energy} or higher entropy scores are identified as OOD nodes.

While these graph OOD detection methods are effective, they invariably rely on the assumption that ground truth ID labels are readily available. This assumption often overlooks a critical challenge: obtaining sufficient high-quality labels for graph-structured data. Specifically, (1) the diverse and complex nature of graph-structured data makes human labeling inherently difficult, and (2) the large scale of real-world graphs renders annotating a significant portion of nodes both time-consuming and resource-intensive \cite{chen2023label}.


\noindent \textbf{Our Observations and Motivation.}
In this paper, we aim to address the challenge of few-shot OOD detection and ID classification on  text-attributed graphs (TAGs) within the commonly used semi-supervised transductive setting, as described above.  
Consider a text-attributed social network where nodes represent individuals, node attributes correspond to their textual descriptions, and edges denote interactions or connections between them. Initially, the entire network is unlabeled, and the goal is to classify individuals into specific interest groups, such as technology enthusiasts, sports fans, or musicians, while operating within a limited human annotation budget. However, the network also contains individuals whose interests fall outside these predefined categories, such as those primarily engaged in political discussions or travel blogging. Identifying and labeling these OOD nodes would be inefficient, as they do not contribute to training an effective classifier for the targeted interest groups. Instead, the focus is on accurately classifying only the ID nodes while detecting and filtering out OOD nodes that do not belong to the intended classification space.
Furthermore, zero-shot \cite{wang2023clipn,ding2024zero} and few-shot \cite{miyai2024locoop,bai2024id} OOD detection for images has been extensively studied using multi-modal foundation models.
However, to date, no similarly powerful graph foundation model exists to support zero-shot or few-shot graph OOD detection. As a result, we turn to LLMs to tackle the data-efficiency challenge of OOD detection on TAGs.

In summary, current graph OOD detection methods typically rely heavily on large amounts of labeled ID nodes to perform well. Conversely, while LLMs demonstrate remarkable zero-shot capabilities on text-attributed graphs (TAGs), they inherently lack the ability to interpret and leverage the structural information essential to TAGs. In this study, we take the first step toward integrating the strengths of both GNNs and LLMs to tackle the data-efficiency challenges in graph OOD detection.

\begin{figure}[!t]
   \begin{center}
   \includegraphics[width=10cm,height=4cm]{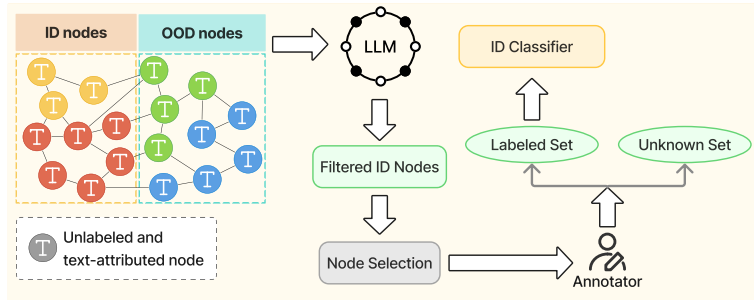}
    \end{center}
 \caption{An illustration of our method. To reduce annotation costs, we use an LLM to filter out OOD nodes before selecting nodes for human annotation. The annotated ID nodes are then used to train the target ID classifier.}
\label{fig:Intro_img}
\end{figure}

\noindent \textbf{Present work.}
As shown in Fig. \ref{fig:Intro_img}, to address these challenges, we propose to leverage LLMs to filter out OOD nodes before human annotation, thereby \textbf{reducing human costs}. Specifically, we provide the LLM with ID knowledge (i.e., the names of ID classes) and prompt it to determine whether an unlabeled query node belongs to one of the ID classes, using the text information associated with the query node. Note that while we enable LLMs to directly perform zero-shot OOD detection and ID classification, using LLMs for zero-shot annotation is very slow during inference. Therefore, we aim to leverage LLMs to reduce human annotation costs during training and rely solely on well-trained GNNs for faster inference during testing.
However, prompting the LLM to annotate all unlabeled nodes in the training set is costly for large graphs, although the cost of using an LLM is significantly lower than that of human annotation. To further \textbf{reduce the LLM’s cost}, we propose prompting the LLM to annotate only a small subset of nodes and then using these pseudo-labels to train a lightweight GNN filter. With this GNN filter, we can predict whether every unlabeled node in the training set belongs to one of the ID classes. If not, it's very likely that this node is an OOD node, and we then filter it out before human annotation.

In addition, we can obtain the embeddings of all unlabeled nodes in the graph after training the GNN filter with pseudo-labels. Based on these embeddings, the most informative nodes can be selected using existing informativeness-aware node selection methods. These selected nodes are then annotated by a human annotator, and the final annotated ID nodes are used to train the target ID classifier. Optionally, we can combine the accurate labels from human annotation with the noisy labels from the LLM to train a robust ID classifier under \textbf{severe data scarcity scenarios}.
Compared to other active learning methods that require multiple rounds of selection \cite{wu2019active,chang2024multitask}, our approach requires only a single round of annotation. Moreover, relying solely on noisy labels from an LLM to train an ID classifier imposes a performance upper bound (see the results in Section \ref{exp:upper}). Furthermore, leveraging LLM knowledge to train smaller models, such as GNNs, facilitates faster inference, particularly in domains where time efficiency is crucial.

\medskip
\noindent
We summarize our key contributions as follows:
\begin{itemize}
\item 
To the best of our knowledge, we are the first to investigate LLM’s zero-shot learning ability for the graph OOD detection problem. 
With the zero-shot learning ability of LLMs, our method achieves high performance with only one round of node selection, compared to traditional multi-round active learning selection methods.

\item We design a general framework LLM-GOOD that can filter out many OOD nodes before annotation to reduce human costs and use LLM’s zero-shot annotations to train a light GNN filter to further reduce LLM costs. 

\item 
We apply \method to node classification datasets consisting of different properties under label budget constraints. 
Experimental results show that our method effectively filters out OOD nodes and achieves much better ID classification and OOD detection performance compared to baselines within an annotation budget. 

\end{itemize}

\section{Related Work}
\subsection{Graph OOD Detection}
In recent years, OOD detection in graph data has presented new challenges, especially in the context of multi-class classification for in-distribution data, which further complicates the task of identifying outlier data \cite{marevisiting}. For instance, OODGAT \cite{song2022learning} leverages a graph neural network (GNN) that explicitly models interactions among different types of nodes, enabling effective separation of inliers and outliers during feature propagation. 
GNNSafe \cite{wu2023energy} highlights the inherent OOD detection capabilities of standard GNN classifiers and proposes a robust OOD discriminator using an energy-based function derived from GNNs trained with standard classification loss. GRASP \cite{marevisiting} explores the potential of OOD score propagation and derives the conditions under which the score propagation is beneficial. They also propose an edge augmentation strategy with theoretical guarantees for post-hoc node-level OOD detection.

While effective, these methods rely heavily on the assumption of abundant ID labels in open-set scenarios. However, in real-world applications, labeled data are costly and challenging to obtain, limiting the practicality of such approaches.

\subsection{Data-Efficient Graph Learning}
Researchers have conducted extensive and focused studies exploring graph machine learning in low-resource settings, with the goal of reducing the costs and time required for annotation \cite{ju2024survey}. Current data-efficient graph learning methods can be broadly divided into three categories: self-supervised graph learning, semi-supervised graph learning, and few-shot graph learning.

Few-shot graph learning aims at enabling models to generalize effectively and make accurate predictions using only a small number of labeled examples. The primary objective is to train models to learn from a limited set of annotated instances and apply this knowledge to predict new and unseen data \cite{ju2024survey}. To achieve this, researchers typically adopt one of two approaches: metric learning, which encourages query nodes to align closely with their respective prototypes \cite{tan2022graph}, or parameter optimization, which employs meta-learning to generate node representations \cite{ju2023few}. Some graph active learning methods \cite{wu2019active,cai2017active,chang2024multitask} have been developed to enhance the performance of semi-supervised node classification while adhering to a label budget constraint. For instance, FeatProp \cite{wu2019active} identifies nodes by propagating their features throughout the graph structure and applying K-Medoids clustering, mitigating the impact of under-trained model representations. However, both current few-shot graph learning methods \cite{ding2020graph,yu2022hybrid} and graph active learning techniques are restricted to the closed-set node classification scenario. Recently, \cite{xu2024lego} applied active learning methods to the graph open-set classification scenario. However, their approach involves using real OOD nodes and requires multiple rounds of node selection for human annotation.

\subsection{LLMs as Prefix for Graphs}
In this paper, we focus on utilizing information generated by LLMs to enhance the training of GNNs. These techniques can be divided into two main categories: 
(i) Embeddings from LLMs for GNNs, which involves incorporating embeddings produced by LLMs into GNNs, and 
(ii) Labels from LLMs for GNNs, which focuses on leveraging labels generated by LLMs to guide GNN training \cite{ren2024survey}. We mainly focus on the second category that leverages generated labels from LLMs as supervision to improve the training of GNNs.

LLM-GNN \cite{chen2023label} utilizes LLMs as annotators to produce node category predictions accompanied by confidence scores, which are treated as labels. A post-filtering process is applied to remove low-quality annotations while ensuring label diversity. These refined labels are then used to train GNNs. Similarly, GraphEdit \cite{guo2024graphedit} uses LLMs to create an edge predictor, which evaluates and refines candidate edges by comparing them to the edges of the original graph.

\section{Setting}
\subsection{Text-Attributed Graphs}
Our study focuses on TAGs, represented as $G_T = (\mathcal{V}, \mathbf{A}, \mathbf{T}, \mathbf{X})$. The set of nodes is $\mathcal{V} = \{v_1, \dots, v_n\}$, where each node is associated with raw text attributes $\mathbf{T} = \{t_1, t_2, \dots, t_n\}$. These text attributes can be converted into sentence embeddings $\mathbf{X} = \{x_1, x_2, \dots, x_n\}$ using SentenceBERT \cite{reimers2019sentence}. The adjacency matrix $\mathbf{A} \in \{0, 1\}^{n \times n}$ encodes graph connectivity, where $\mathbf{A}[i, j] = 1$ indicates an edge between nodes $i$ and $j$. 

\subsection{Graph OOD Detection}
\label{subsec:prob_def}
The node set can be partitioned as $\mathcal{V} = \mathcal{V}_\text{in} \cup \mathcal{V}_\text{out}$, where $\mathcal{V}_\text{in}$ denotes the set of ID nodes, and $\mathcal{V}_\text{out}$ represents the set of OOD nodes. We assume that ID nodes are drawn from the distribution $P^{\text{in}}_{\mathcal{V}}$, while OOD nodes are sampled from the distribution $P^{\text{out}}_{\mathcal{V}}$. The OOD node detection task is formally defined as follows: Given a collection of nodes sampled from $P^{\text{in}}_{\mathcal{V}}$ and $P^{\text{out}}_{\mathcal{V}}$, the objective is to accurately determine the source distribution—either $P^{\text{in}}_{\mathcal{V}}$ or $P^{\text{out}}_{\mathcal{V}}$—for each node.  

We study OOD node detection in graphs under the transductive learning paradigm, where ID and OOD nodes coexist in the same graph, the most common framework for node-level OOD detection. During training, only the node attributes $\mathbf{X}$, the adjacency matrix $\mathbf{A}$, and the ID labels of a subset of nodes, $\mathcal{V}' \subseteq \mathcal{V}_\text{in}$, are provided. In general, the task consists of two main objectives: (1) \textbf{OOD Detection}: For each node $v \in \mathcal{V}$, determine whether it belongs to one of the ID known classes or to an OOD unknown class. (2) \textbf{ID Classification}: For nodes identified as ID, assign them to one of the predefined $K$ classes.

\subsection{Few-Shot Graph OOD Detection}
Assume that we have a validation set $\mathcal{V}_{val}$ and a test set $\mathcal{V}_{test}$. The remaining nodes form the candidate set $\mathcal{V}_{can} = \mathcal{V} \setminus (\mathcal{V}_{val} \cup \mathcal{V}_{test})$. All nodes in $\mathcal{V}_{can}$ are initially \textbf{unlabeled}. Given a human label budget $\mathcal{B}$, our goal is to select a subset of nodes from $\mathcal{V}_{can}$ such that the trained model $f$ achieves the lowest expected loss in the test set $\mathcal{V}_{test}$:

\begin{equation}
    \arg\min_{\mathcal{V}_{can}^{s} \subset \mathcal{V}_{can}, |\mathcal{V}_{can}^{s}| = \mathcal{B}}
    \mathbb{E}_{v_i \in \mathcal{V}_{test}} \left[ \ell(y_i, \tilde{y}_i) \right]
\end{equation}

where $f$ is our target ID classifier, ${y}_i$ is the ground truth label of node $v_i$, and $\tilde{y}_i$ denotes the label prediction of node $v_i$ by $f$. Compared with other label-efficient graph learning methods, such as active learning approaches, \textbf{we do not require any initial set of labeled nodes and, more importantly, we only select nodes for one round of annotation}.

\section{Method}

\begin{figure*}[!t]
   \begin{center}
   \includegraphics[width=1\linewidth]{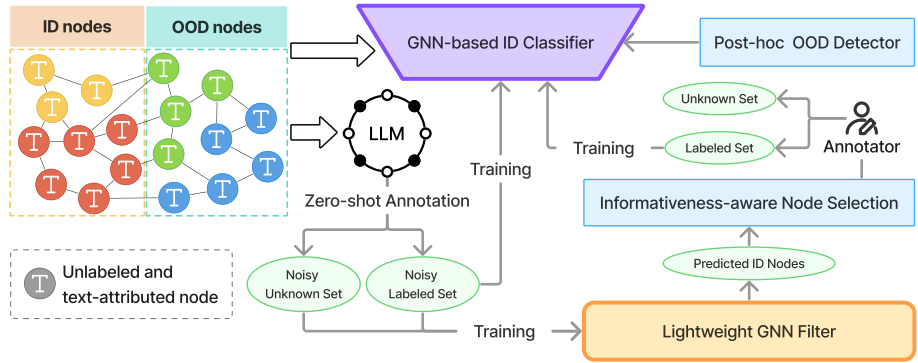} %
 \vspace{-1.0cm}
    \end{center}
 \caption{An overview of our framework \method. 
To reduce human cost, we use LLM to filter out OOD nodes before human annotation (\S \ref{subsec:LLM}). To further reduce LLM cost, we use LLM to annotate a small subset of nodes, and then train a lightweight GNN filter on these noisy annotations to predict labels for the remaining nodes in the graph (\S \ref{subsec:filter}). 
After obtaining node embeddings from the GNN filter, informativeness-aware selection methods identify the most informative unlabeled potential ID nodes (\S \ref{subsec:selection}). After these selected nodes are annotated, the labeled accurate ID nodes are used to train the target ID classifier for ID classification and OOD detection (\S \ref{subsec:classifier}).
}
\label{fig:framwork_pdf_1009}
\end{figure*}

In real-world scenarios, graphs typically include a large number of unlabeled nodes, many of which may be OOD nodes and irrelevant to the target task. Our goal is to train an ID classifier using a limited set of ID labels, striving for high accuracy in ID classification while effectively identifying OOD data, where the classifier should exhibit low confidence.

To reduce human efforts, we seek to exclude as many OOD nodes as possible from the training set prior to labeling. To achieve this, the first step is to use an LLM as an annotator to identify potential OOD nodes (see \S \ref{subsec:LLM}). However, annotating all unlabeled nodes in the training set using the LLM still incurs a high cost. Therefore, we propose to annotate a small subset of nodes with the LLM and use these pseudo-labels from LLM to train a lightweight GNN filter (see \S \ref{subsec:filter}). This approach further reduces the cost of using the LLM. After training, the GNN filter can predict which unlabeled nodes are ID nodes, allowing us to identify potential ID nodes with minimal use of the LLM.

Furthermore, based on the node embeddings from the GNN filter, informativeness-aware node selection methods, such as K-Medoids-based node selection, can be applied to choose the most informative nodes from the unlabeled potential ID nodes (see \S \ref{subsec:selection}). Once these informative nodes are annotated, the labeled ID nodes can be used to train the target ID classifier.
Optionally, accurate labels from humans and noisy labels from the LLM can be combined to train a robust ID classifier, especially in scenarios of extreme data scarcity.
Finally, post-hoc OOD detection methods can be applied to the classifier to enhance its ability to recognize unseen classes (see \S \ref{subsec:classifier}).
Fig. \ref{fig:framwork_pdf_1009} illustrates the pipeline of the proposed framework \method. 

\subsection{LLM as Zero-shot Open-world Annotator}
\label{subsec:LLM}
We randomly select a small set of nodes $\mathcal{V}_{LLM}$ from $\mathcal{V}_{can}$ and then let LLM annotate them.
We provide the LLM with ID knowledge (ID classes' names) and prompt it to determine whether an unlabeled query node belongs to one of the ID classes, incorporating the text information of the query node. An example prompt for zero-shot OOD detection is shown in the following box.
We instruct LLM to output "none" if it predicts that the node does not belong to any of the ID classes. 
Therefore, the noisy labels of $\mathcal{V}_{LLM}$ from LLM is $\mathcal{Y}_{LLM} = \{y_1^n, y_2^n, \dots, y_m^n\}$, where $m$ is the number of annotated nodes. Given $K$ known ID classes, the LLM's label set extends to $K+1$ classes, with the $(K+1)$-th class representing the unknown class.

\begin{tcolorbox}
[colback=green!10,colframe=gray,fonttitle=\bfseries, title=Zero-Shot OOD Detection and ID Classification Prompt, fonttitle=\bfseries]
\label{box:annotation_short_prompt}

As a research scientist, your task is to analyze and classify \textbf{\{object\}} based on their main topics, meanings, background, and methods. Please first read the content of the \textbf{\{object\}} carefully. Then, identify the \textbf{\{object\}}'s key focus. Finally, match the content to one of the given categories.

There are the following categories:
\texttt{[Category 1, Category 2, Category 3, ...]}

Given the current possible categories, determine if it belongs to one of them. If so, specify that category; otherwise, say \texttt{"none"}.


\texttt{[Insert \textbf{\{Object\}} Content Here]}
\end{tcolorbox}



\subsection{Train Lightweight GNN with Pseudo-Labels}
\label{subsec:filter} 
To further reduce LLM's cost, we first use the LLM to annotate a small subset of nodes and then train a GNN on these annotations to predict labels for the remaining nodes in the graph. With the labeled node set $\mathcal{V}_{LLM}$ and its noisy labels $\mathcal{Y}_{LLM}$, we can train a $K+1$ class classifier.
As an aside, any GNN can be used as the lightweight OOD filter. In this paper, we use a two-layer standard graph convolutional network (GCN) as the OOD filter, and set the output dimension of the last layer as ${K+1}$. The output of the first layer is as follows:
\begin{equation}
    \mathbf{H}^{(1)} = \sigma \left( \tilde{\mathbf{D}}^{-\frac{1}{2}} \tilde{\mathbf{A}} \tilde{\mathbf{D}}^{-\frac{1}{2}} \mathbf{X} \mathbf{W}^{(0)} \right)
\end{equation}
where $\tilde{\mathbf{A}} = \mathbf{A} + \mathbf{I}$ and $\tilde{\mathbf{D}}_{ii} = \sum_j \tilde{\mathbf{A}}_{ij}$, $\mathbf{I}$ is the identity matrix, and $\mathbf{W}^{(0)}$ is the weight matrix. 
The OOD filter's final output for all nodes is $\mathbf{H}^{(2)} \in \mathbb{R}^{N\times (K+1)}$:  
\begin{equation}
    \mathbf{H}^{(2)} = \sigma \left( \tilde{\mathbf{D}}^{-\frac{1}{2}} \tilde{\mathbf{A}} \tilde{\mathbf{D}}^{-\frac{1}{2}} \mathbf{H}^{(1)} \mathbf{W}^{(1)} \right)
\end{equation}

Embeddings $\mathbf{H^{(1)}}$ capture graph structure information and can be leveraged in the subsequent module for selecting nodes based on informativeness. In addition, with $\mathbf{H}^{(2)}$, we can determine whether each unlabeled node belongs to the unknown $(K+1)$-th class, with the goal of filtering out as many OOD nodes as possible prior to human annotation. As a result, we retain nodes predicted to belong to one of the first $K$ ID classes for further processing, while excluding those identified as unknown. Specifically, our goal is to filter out OOD nodes from $\mathcal{V}_{can}$ based on $\mathbf{H}^{(2)}$ to get the filtered ID node set $\mathcal{V}_{can}^{ID}$ and then select the most informative nodes from $\mathcal{V}_{can}^{ID}$ based on $\mathbf{H}^{(1)}$.

\begin{equation}
    \mathcal{V}_{can}^{ID} = \left\{v_i,   \operatorname*{arg\,max}_{k} \mathbf{H}^{(2)}[i, K] \leq K \right\}
\end{equation}

The cross-entropy loss function of the OOD filter is defined as:

\begin{equation}
    \mathcal{L} = -\frac{1}{|\mathcal{V}_{LLM}|} \sum_{i \in \mathcal{V}_{LLM}} \sum_{k=1}^{K+1} y_{ik}^{n} \log \hat{y}_{ik}^{n}
    \label{equation::classifier}
\end{equation}

\subsection{Informativeness-aware Node Selection}
\label{subsec:selection}
Most node selection methods typically prioritize nodes with high prediction uncertainty or diverse representations for labeling. However, in the presence of open-set noise, these metrics become unreliable, as OOD nodes also exhibit high uncertainty and diversity while lacking class-specific features or shared inductive biases with ID examples. By utilizing our OOD filter to remove a significant number of OOD nodes, we can more effectively identify and select the most informative nodes from the remaining potential ID nodes. Any graph active selection method, such as FeatProp \cite{wu2019active} or MITIGATE \cite{chang2024multitask}, can be applied.

\subsection{ID Classification and OOD Detection}
\label{subsec:classifier}
With the help of the OOD filter, we can train the target ID classifier with more labeled ID nodes while adhering to the label budget constraint.


Assume that we have selected $\mathcal{V}_{can}^s$ from $\mathcal{V}_{can}^{ID}$ and annotated it with accurate labels $\mathcal{Y}_{can}^s$. We now have a set of nodes, $\mathcal{V}_{can}^s$, with accurate labels $\mathcal{Y}_{can}^s$, and a set of nodes, $\mathcal{V}_{LLM}$, with noisy labels $\mathcal{Y}_{LLM}$. From $\mathcal{V}_{can}^s$ and $\mathcal{V}_{LLM}$, we can derive the ID node set $\mathcal{V}_{can}^{s-ID}$ with accurate labels and $\mathcal{V}_{LLM}^{ID}$ with noisy labels. We can then use $\mathcal{V}_{can}^{s-ID}$ and $\mathcal{V}_{LLM}^{ID}$ to train the target ID classifier. Similarly, any graph neural network can serve as the ID classifier. The design of noise-resistant GNNs to better leverage the noisy labels from LLM is left for future study.

Specifically, the output of the ID classifier is $\mathbf{Z} \in \mathbb{R}^{N\times K}$:
\begin{equation}
    \mathbf{Z} = GNN(\mathbf{A},\mathbf{X})
\end{equation}

Note that if a node is in both $\mathcal{V}_{can}^{s-ID}$ and $\mathcal{V}_{LLM}^{ID}$, its label is taken from $\mathcal{Y}_{can}^{s-ID}$. Using noisy labels from $\mathcal{Y}_{LLM}^{ID}$ is extremely helpful when there are very few accurate labels available, particularly in situations of extreme data scarcity.

After training the ID classifier, any post-hoc OOD detector \cite{liang2017enhancing,lee2018simple,hendrycks2016baseline,yang2022openood,marevisiting} can be applied to the output logits of the ID classifier. As an example, consider the well-known post-hoc OOD detector, MSP \cite{hendrycks2016baseline}. Correctly classified examples generally exhibit higher maximum softmax probabilities compared to misclassified and out-of-distribution examples. Consequently, given $\mathbf{Z}$, we can compute the softmax probability of the predicted class, i.e., the maximum softmax probability, which serves as the OOD score.

\section{Experiments}
Our experiments answer the following research questions (RQ): RQ1 (\S \ref{exp:Main Results}): How effective is the proposed \method in ID classification and OOD detection compared to other leading baselines? RQ2 (\S \ref{exp:Main Results}): Whether LLMs can filter out OOD nodes effectively? RQ3 (\S \ref{exp:scarcity}): Will \method be robust to different settings, such as varying levels of label scarcity? RQ4 (\S \ref{exp:LLMs}): What are the differences in cost and effectiveness between various LLMs?   

\subsection{Experimental Setup}
\subsubsection{Datasets}
We utilize the following TAG datasets, which are commonly used for node classification: Cora \cite{mccallum2000automating}, Citeseer \cite{giles1998citeseer}, Pubmed \cite{sen2008collective} and Wiki-CS \cite{mernyei2020wiki}.
For each dataset, we split all classes into ID and OOD sets, and the ID classes for the four datasets are shown in Appendix B.
Additionally, the number of ID classes is set to a minimum of two to perform the ID classification task.

For each dataset with $K$ ID classes, we randomly select $10\times K$ ID nodes and an equal number of OOD nodes for validation. The test set consists of 500 ID and 500 OOD nodes, while the remaining nodes form $\mathcal{V}_{can}$.

\subsubsection{Baselines}
We evaluate \method against two categories of baselines: (1) OOD detection methods, including MSP \cite{hendrycks2016baseline}, Entropy, GNNSafe \cite{wu2023energy}, and GRASP \cite{marevisiting}; (2) node selection methods for node classification, including uncertainty-based selection \cite{luo2013latent}, FeatProp \cite{wu2019active}, and MITIGATE \cite{chang2024multitask}, where different selection strategies are integrated into GCNs with MSP as the OOD score.


For all methods, including baselines and \method, we use two GCN layers as the ID classifier. 

\subsubsection{Settings}
For all datasets, we use GPT-4o-mini to annotate 200 randomly selected nodes and train the lightweight GNN filter using these annotated noisy nodes with two standard GCN layers. The results for other LLM are given in Section \ref{exp:LLMs}.
For \method, we use the energy score \cite{wu2023energy} as the OOD score. 
Additionally, we evaluate an alternative approach (\method-f), where LLMs filter all unlabeled nodes in the initial graph, and a subset of ID-labeled nodes is randomly selected for manual labeling.

\subsubsection{Evaluation Metrics}
For the ID classification task, we use classification accuracy (ID ACC) as the evaluation metric. For the OOD detection task, we employ three commonly used metrics from the OOD detection literature \cite{song2022learning}: the area under the ROC curve (AUROC), the precision-recall curve (AUPR), and the false positive rate when the true positive rate reaches 95\% (FPR@95). In all experiments, the OOD nodes are considered positive cases. Details about these metrics are provided in Appendix A.

\subsubsection{Implementation Details}
We evaluate all methods under the total label budgets $10\times K$ and $5\times K$, respectively. Since baseline methods require an initial set of labeled nodes and multiple rounds of node selection, in each selection round, $K$ nodes are chosen from the unlabeled pool and annotated for all baselines. In addition, we allocate an initial label budget of $5\times K$ for the total budget of $10\times K$ and $K$ for the total budget of $5\times K$. In contrast, our method does not require an initial set of labeled nodes and involves only a single round of random node selection for annotation.

All GCNs have 2 layers with hidden dimensions of 32. All models use a learning rate of 0.01, a dropout probability of 0.5 and a weight decay of 5\text{e-}4. For all K-Medoids-based selection methods, the number of clusters is fixed at 48. For \method, 200 nodes are randomly selected and annotated by the LLM. The weight assigned to the unknown class in the GNN filter's loss function is selected from $\{0.05, 0.1, 0.2, 0.3, 0.5\}$ based on the performance of the validation set. For all experiments, we average all results across 5 different random seeds.

\begin{table*}[!t]
\centering
\renewcommand{\arraystretch}{1.6} 
\setlength{\tabcolsep}{5.9pt} 
\fontsize{24pt}{24pt}\selectfont 

\caption{Performance comparison (best highlighted in bold) of different models on ID classification and OOD detection tasks for the Cora and Citeseer datasets under label budget $10\times K$. All values are percentages (\%).}
\label{table:1a}
\begin{adjustbox}{max width=\textwidth} 

\begin{tabular}{c|cccc|cccc}
\hline
\multirow{1}{*}{\textbf{Model}}  & \multicolumn{4}{c|}{\textbf{Cora}} & \multicolumn{4}{c}{\textbf{Citeseer}} \\ 
                                 &\textbf{ID ACC} $\uparrow$& \textbf{AUROC} $\uparrow$& \textbf{AUPR} $\uparrow$& \textbf{FPR@95} $\downarrow$
                                 & \textbf{ID ACC} $\uparrow$& \textbf{AUROC} $\uparrow$ & \textbf{AUPR} $\uparrow$& \textbf{FPR@95} $\downarrow$ \\ \hline
 
GCN-Uncertainty &  79.04$\pm$7.98 &  77.02$\pm$4.46 &   79.51$\pm$3.61 &  75.80$\pm$7.17  
                &  75.36$\pm$4.81 &  69.73$\pm$5.08 &    69.57$\pm$5.58 &   87.72$\pm$4.93 \\ 

GCN-FeatProp   &  81.04$\pm$2.45 &  78.24$\pm$3.25 &   79.92$\pm$4.07 & 75.92$\pm$5.40  
               &  79.48$\pm$2.83 & 71.45$\pm$4.47  &  71.42$\pm$4.60 &    86.56$\pm$6.41 \\ 

GCN-MITIGATE   &  81.64$\pm$2.31 &  79.04$\pm$2.31 &  80.52$\pm$2.56 &  73.40$\pm$4.71  
               &  80.44$\pm$3.12 &  72.19$\pm$4.33 &  71.92$\pm$3.99 &  \textbf{84.52$\pm$6.08} \\ \hline

MSP           &  77.68$\pm$7.60 &  75.40$\pm$6.85 &  78.19$\pm$5.53 &  81.32$\pm$9.72  
              &  70.92$\pm$7.46 &  62.12$\pm$7.09 & 64.63$\pm$5.02 &  90.64$\pm$3.78 \\ 

GNNSafe       &  74.76$\pm$8.99 &  84.05$\pm$7.44 & 84.62$\pm$6.42 &  61.20$\pm$19.24  
              &  71.16$\pm$7.44 &  65.84$\pm$5.73 & 65.97$\pm$5.13 &  89.12$\pm$3.29 \\ 

Entropy       &  76.80$\pm$8.65 &  76.10$\pm$8.08 &  78.12$\pm$6.68 &  76.24$\pm$12.87  
              &  73.20$\pm$4.28 &  63.26$\pm$6.65 &  65.24$\pm$4.81 &  88.56$\pm$4.69 \\ 

GRASP         &  77.88$\pm$8.36 &  83.00$\pm$6.43 &  82.30$\pm$6.33 &  61.48$\pm$21.59  
              &  71.72$\pm$5.37 &  60.64$\pm$6.62 &  63.04$\pm$4.67 &  91.20$\pm$2.58 \\ \hline


LLM-GOOD-f    &  84.00$\pm$4.40 & 86.59$\pm$2.32 & 87.36$\pm$3.10 & 60.56$\pm$3.94
              &  72.52$\pm$10.43 & 70.71$\pm$4.49 & 72.99$\pm$4.77 & 88.92$\pm$6.85 \\

\method       & \textbf{85.20$\pm$2.68}  & \textbf{88.06$\pm$3.77} & \textbf{87.85$\pm$3.68}  & \textbf{48.04$\pm$1.19} 
              & \textbf{80.60$\pm$3.38} &\textbf{73.29$\pm$4.12} & \textbf{75.26$\pm$3.34} &  86.48$\pm$5.45 \\ 

\hline
\end{tabular}
\end{adjustbox}
\end{table*}

\begin{table*}
\centering
\renewcommand{\arraystretch}{1.6} 
\setlength{\tabcolsep}{5.9pt} 
\fontsize{24pt}{24pt}\selectfont 

\caption{Performance comparison (best highlighted in bold) of different models on ID classification and OOD detection tasks for the Pubmed and Wiki-CS datasets under label budget $10\times K$. All values are percentages (\%).}
\label{table:1b}
\begin{adjustbox}{max width=\textwidth} 

\begin{tabular}{c|cccc|cccc}
\hline
\multirow{1}{*}{\textbf{Model}}  & \multicolumn{4}{c|}{\textbf{Pubmed}} & \multicolumn{4}{c}{\textbf{Wiki-CS}} \\ 
                                 &\textbf{ID ACC} $\uparrow$& \textbf{AUROC} $\uparrow$& \textbf{AUPR} $\uparrow$& \textbf{FPR@95} $\downarrow$
                                 & \textbf{ID ACC} $\uparrow$& \textbf{AUROC} $\uparrow$ & \textbf{AUPR} $\uparrow$& \textbf{FPR@95} $\downarrow$ \\ \hline
 
GCN-Uncertainty &  84.48$\pm$9.41 &  57.15$\pm$6.61 &  55.96$\pm$5.69 & 91.60$\pm$2.39 
                & 81.88$\pm$4.70   &  77.31$\pm$6.18  &   79.59$\pm$5.82&  78.48$\pm$10.79\\ 

GCN-FeatProp   &  83.00$\pm$8.57 &  53.07$\pm$11.45 & 53.58$\pm$9.42 & 92.96$\pm$5.64   
               &  76.48$\pm$5.49 & 73.58$\pm$6.99 &75.26$\pm$7.90  & 83.52$\pm$6.43 \\ 

GCN-MITIGATE   &  83.24$\pm$8.48 &  57.91$\pm$10.41 &  57.93$\pm$9.57 & 92.60$\pm$3.76  
               &  77.76$\pm$7.96   & 71.69$\pm$4.91   &  73.26$\pm$4.77& 86.16$\pm$7.90 \\ \hline

MSP           &  81.04$\pm$7.51  &  52.65$\pm$8.86  &  53.74$\pm$7.15  &  93.44$\pm$3.60
              &  77.52$\pm$5.60  &  75.10$\pm$4.56  &  77.54$\pm$5.48  &  84.80$\pm$4.72 \\ 

GNNSafe       &  82.16$\pm$8.02  &  49.65$\pm$17.49  &  55.55$\pm$14.82  &  93.04$\pm$6.90  
              &  78.80$\pm$4.72  &  83.71$\pm$4.38  &  84.95$\pm$06.08  &  82.16$\pm$11.53 \\ 

Entropy       &  81.36$\pm$7.57  &  52.22$\pm$09.70  &  52.61$\pm$8.47  &  92.68$\pm$3.27 
              &  77.76$\pm$4.80  &  72.85$\pm$3.71  &  75.47$\pm$4.68  &  88.00$\pm$3.46 \\ 

GRASP         &  82.68$\pm$8.18  &  49.97$\pm$19.59  &  54.86$\pm$14.99  &  93.76$\pm$5.46 
              &  78.28$\pm$5.11  &  78.34$\pm$8.54  &  79.17$\pm$10.88  &  86.28$\pm$6.14 \\ \hline


LLM-GOOD-f    &  87.00$\pm$2.19  & 61.09$\pm$19.05  & 66.81$\pm$15.58  & 91.24$\pm$4.89  
              &  83.76$\pm$2.46  & 86.84$\pm$1.84  & 89.42$\pm$1.57  & 79.24$\pm$10.66 \\

\method       & \textbf{87.08$\pm$2.58}  & \textbf{64.87$\pm$16.70}  & \textbf{70.60$\pm$14.26}  & \textbf{90.72$\pm$5.10}  
              & \textbf{83.92$\pm$3.53}  & \textbf{87.71$\pm$2.41}  & \textbf{89.84$\pm$2.64}  & \textbf{71.04$\pm$16.20} \\ 

\hline
\end{tabular}
\end{adjustbox}
\end{table*}

\subsection{Main Results}
\label{exp:Main Results}
As shown in Tables \ref{table:1a}, \ref{table:1b} and \ref{table:2}, \method consistently outperforms state-of-the-art graph OOD detection methods by a significant margin across all TAG datasets.  
Specifically, for ID classification on four datasets, the most substantial improvement is observed on the Cora dataset when the label budget is set to $5 \times K$. In this case, the ID accuracy increases from 63.80\% (achieved by the best baseline, GNNSafe) to 81.52\%, reflecting a notable improvement of 17.72\%.  
It is important to note that all baselines have an initial set of labeled ID nodes and use multiple selection rounds to improve performance. In contrast, \method selects nodes randomly in a single round yet still outperforms the baselines.

Furthermore, \method exhibits remarkable advancements in OOD detection metrics, achieving higher AUROC and AUPR scores while maintaining a lower FPR@95 across all datasets. The most significant improvement is observed in the Pubmed dataset when the label budget is $10\times K$, where the AUROC increases from 57.91\% (achieved by the best baseline, GCN with MITIGATE node selection) to 64.87\%, marking an improvement of 6.96\%.

Moreover, we calculate the final proportion of ID nodes, that is, the ratio of true ID nodes to the total number of selected and annotated nodes, across various selection methods. 
The results are in Appendix C.
Our method achieves the highest proportion across all datasets compared to the baselines. This shows that our method effectively filters out OOD nodes before human annotation, thereby reducing annotation costs.
While \method and \method-f achieve similar ID node proportions, \method significantly reduces LLM costs by annotating only a small number of nodes and leveraging a GNN filter to label the rest.

\begin{table*}[!t]
\centering
\renewcommand{\arraystretch}{1.6} 
\setlength{\tabcolsep}{5.9pt} 
\fontsize{24pt}{24pt}\selectfont 

\caption{Performance comparison (best highlighted in bold) of different models on ID classification and OOD detection for four TAG datasets under label budget $5\times K$. Our method achieves the best across all baselines. All values are percentages (\%).}
\label{table:2}
\begin{adjustbox}{max width=\textwidth} 

\begin{tabular}{c|cccc|cccc|cccc|cccc}
\hline
\multirow{1}{*}{\textbf{Model}}  & \multicolumn{4}{c|}{\textbf{Cora}} & \multicolumn{4}{c|}{\textbf{Citeseer}} & \multicolumn{4}{c|}{\textbf{Pubmed}} & \multicolumn{4}{c}{\textbf{Wiki-CS}} \\ 
                                 &\textbf{ID ACC} $\uparrow$& \textbf{AUROC} $\uparrow$& \textbf{AUPR} $\uparrow$& \textbf{FPR@95} $\downarrow$
                                 & \textbf{ID ACC} $\uparrow$& \textbf{AUROC}$\uparrow$ & \textbf{AUPR} $\uparrow$& \textbf{FPR@95}$\downarrow$ 
                                 &\textbf{ID ACC} $\uparrow$& \textbf{AUROC} $\uparrow$& \textbf{AUPR} $\uparrow$& \textbf{FPR@95} $\downarrow$
                                 &\textbf{ID ACC} $\uparrow$& \textbf{AUROC} $\uparrow$& \textbf{AUPR} $\uparrow$& \textbf{FPR@95} $\downarrow$\\ \hline

GCN-Uncertainty &  51.20  &  65.51  &  65.90  &  87.88  &  59.96  &  67.53  &  69.00  &  89.76  &  73.36  &  57.53  &  57.55  &  90.56  &61.40&61.94&64.86  & 92.12\\ 
GCN-FeatProp &  55.76  &  71.00  &  72.28  &  85.88  &  69.32  &  67.00  &  65.87  &  \textbf{86.20}  &  71.60  &  50.89  &  51.19  &  95.04  & 71.64 & 69.30 & 71.73 & 90.32\\ 
GCN-MITIGATE &  58.68  &  67.36  &  69.36  &  86.64  &  67.64  &  64.42  &  65.27  &  90.72  &  71.76  &  57.27  &  57.73  &  92.84  &70.36 & 60.36 & 62.29 & 92.84 \\ \hline
MSP &  63.32  &  72.18  &  73.41  &  82.92  &  71.20  &  63.59  &  65.51  &  89.72  &  82.04  &  57.10  &  57.52  &  92.32  &  67.60  &  74.64  &  77.92  &  84.16 \\ 
GNNSafe &  62.52  &  79.29  &  81.12  &  68.76  &  70.88  &  67.47  &  67.16  &  87.52  &  81.48  &  53.71  &  59.51  &  96.44  &  68.20  &  79.38  &  81.26  &  83.28 \\ 
Entropy &  63.80  &  73.03  &  73.06  &  77.08  &  69.00  &  65.47  &  67.14  &  87.20  &  82.04  &  56.80  &  55.75  &  92.16  &  66.40  &  72.45  &  74.96  &  84.04 \\
GRASP &  63.52  &  75.30  &  75.07  &  68.96  &  72.56  &  60.64  &  62.40  &  91.12  &  81.40  &  52.35  &  57.30  &  94.96  &  67.84  &  77.54  &  79.95  &  85.60 \\ \hline
LLM-GOOD-f  & \textbf{81.52} & \textbf{82.26} & \textbf{83.71}  & \textbf{64.72} & 69.20  & \textbf{70.99}  & \textbf{73.39}  & 88.60  & \textbf{83.52} & \textbf{63.29} & \textbf{68.68} & 93.32 & 75.24  & \textbf{84.62}  & \textbf{86.30}  & \textbf{77.88} \\
\method  & 78.60 & 80.21 & 80.82 & 68.88  & \textbf{72.12}  & 67.81  & 69.65  & 91.92  & 78.08 & 60.34 & 65.13 & \textbf{89.76} & \textbf{78.56} & 83.35 & 86.26 & 84.84 \\ \hline
\end{tabular}
\end{adjustbox}
\end{table*}

\subsection{OOD Detection Performance Upper Bound of LLM's m}
\begin{figure}[!t]
   \begin{center}
   \includegraphics[width=1\linewidth,keepaspectratio]{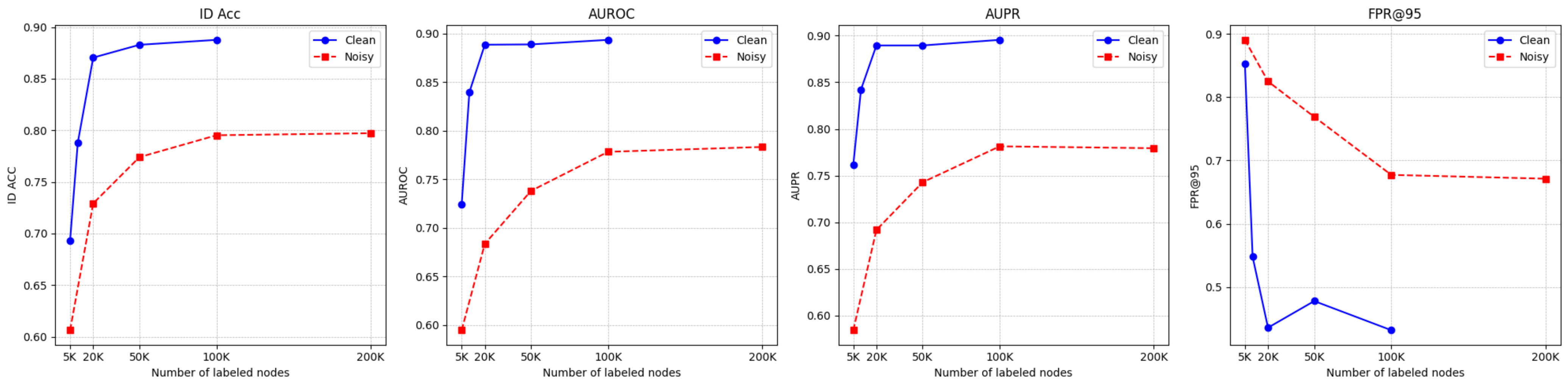}
    \end{center}
 \caption{ID classification and OOD detection Performance upper bound.}
\label{fig:clean_noisy}
\end{figure}

We use different number of LLM's noisy labels and human's annotated accurate labels to train the ID classifier respectively. Given the different label budgets, we randomly select a corresponding number of nodes and use the ID nodes from the selected nodes to train the ID classifier. The results in Figure \ref{fig:clean_noisy} shows that: 
\begin{itemize}
    \item When the number of noisy ID labels or accurate ID labels increases, both ID classification and OOD detection performance improve. However, the improvement rate is significantly higher when using accurate ID labels.
    \item When training the ID classifier with LLM-generated noisy labels, both ID classification and OOD detection performance reach an upper bound substantially lower than that of training with accurate labels.  This highlights the importance of our method, which utilizes LLM to reduce human annotation costs without relying entirely on the LLM for OOD detection.
    \item When the label budget for accurate labels reaches $10 \times K$, ID classification and OOD detection performance nearly reach the upper bound achieved with a large number of noisy labels. At $20 \times K$, both exceed the upper bound of using any number of LLM-generated noisy labels.
\end{itemize}
\label{exp:upper}

\subsection{Combine accurate labels and noisy labels}
\label{exp:scarcity}
We test different methods' performance under severe data-scarcity situation on Cora. The human label budget is set to $1 \times K$, $2 \times K$ and $3 \times K$. For \method-combined, we use 100 noisy labels along with a small number of corresponding clean labels to train the ID classifier. The results are shown in Table \ref{Table:scarcity}.
\begin{table}[H]
\caption{Different methods’ performance under severe data-scarcity situation on Cora. \method-combined achieves the best performance.}
\label{Table:scarcity}
\centering
\begin{adjustbox}{max width=\linewidth}
\begin{tabular}{lccc}
\hline
             & \textbf{$1\times K$} & \textbf{$2\times K$} & \textbf{$3\times K$}\\ \hline
\textbf{GCN-MSP}  &     0.3228       &   0.4260     &  0.4856      \\ 

\textbf{\method} &       0.4580     &     0.6156     &     0.7492          \\
\textbf{\method-combined}         & \textbf{0.7832}          & \textbf{0.8072}            & \textbf{0.8164}         \\ \hline
\end{tabular}
\end{adjustbox}
\end{table}

From the results, we observe that for all methods, increasing the label budget leads to improved performance, and our method consistently outperforms the baseline. When the accurate label budget is extremely small, incorporating additional noisy labels is particularly beneficial. For instance, when the accurate label budget is $1\times K$, the performance gap between \method and \method-combined is $32.52\%$. However, as the accurate label budget increases to  $3\times K$, the performance gap decreases to $6.72\%$. 

Currently, all graph machine learning research assumes either that the entire training set is clean or that all training labels are uniformly affected by a specific type of noise. However, in real-world scenarios, it is more likely that a graph contains a small set of clean labels alongside another set of noisy labels. We leave the design of a more effective pipeline to leverage both label sets to denoise and train a robust, noise-resistant GNN as a direction for future study.

\subsection{LLMs as Open-world Zero-shot Annotators}
We record the annotation cost and zero-shot OOD detection performance of the following LLMs: GPT-3.5-turbo, GPT-4, GPT-4o, GPT-4o-mini, DeepSeek-V3, DeepSeek-R1. Severe rate limitation prevented DeepSeek-R1 from annotating 200 nodes in a reasonable time, so its results are not included.
\label{exp:LLMs}

\subsubsection{Zero-shot annotation accuracy}
We use the baseline short prompt (as shown in Section \ref{subsec:LLM}) and the long prompt (as shown in Appendix D) for zero-shot OOD detection on the Cora dataset. We randomly select 200 nodes and have different LLMs perform zero-shot open-world annotation using these two prompts. The true OOD proportion of the selected nodes is 56\%. The OOD detection performance and the LLMs' predicted OOD proportions are presented in Table \ref{Table:prompt_comparison}. From the results, we can observe that, sometimes, GPT-3.5-turbo does not dare to say 'none', but our long prompt mitigates this issue.
Additionally, both the OOD detection performance and the predicted OOD proportion improve significantly with the long prompt, suggesting that GPT-3.5-turbo becomes more willing to say 'none.' Furthermore, when using the same prompt for open-world annotation, GPT-4o-mini generally outperforms GPT-3.5-turbo in OOD detection.

\begin{table*}[t]
\caption{Comparison of zero-shot OOD detection performance using different prompts across various LLMs on the Cora and Pubmed datasets.}
\label{Table:prompt_comparison}
\centering
\begin{adjustbox}{max width=\textwidth}
\begin{tabular}{lccc|ccc}
\hline
             & \multicolumn{3}{c|}{\textbf{Cora}} & \multicolumn{3}{c}{\textbf{Pubmed}} \\ 
             & \textbf{AUROC} & \textbf{AUPR} & \textbf{OOD Proportion} 
             & \textbf{AUROC} & \textbf{AUPR} & \textbf{OOD Proportion} \\ \hline

\textbf{GPT-3.5-turbo-short prompt}  &  0.5077  &   0.6609   &    0.0200    
                                     &  0.5000  &   0.7100   &    0.0000    \\ 

\textbf{GPT-3.5-turbo-long prompt}   &  0.5909  &   0.7468   &    0.1700    
                                     &  0.5255  &   0.6440   &    0.0300    \\

\textbf{GPT-4o-mini-short prompt}    &  0.7159  &   0.8200   &  0.5600      
                                     &  0.5060  &   0.7135   &  0.0050      \\

\textbf{GPT-4o-mini-long prompt}     & \textbf{0.7366}  & \textbf{0.8323}  & 0.5150      
                                     &  0.8524  & 0.8796    & 0.3650       \\ 

\textbf{ds-v3-short prompt}         & 0.6185  & 0.7589    & 0.2350        
                                    & 0.5000  & 0.7100    & 0.0000        \\

\textbf{ds-v3-long prompt}         & 0.6887  & 0.8170    & 0.7000        
                                   & \textbf{0.9364}  & \textbf{0.9293}  & 0.4700        \\



\hline
\end{tabular}
\end{adjustbox}
\end{table*}

We further evaluate open-world annotation on the PubMed dataset by randomly selecting 200 nodes and having the LLMs annotate them using two prompts. The true OOD proportion of the selected nodes is 42\%. We can observe that our proposed prompt outperforms the baseline short prompt in zero-shot OOD detection, even though the latter explicitly instructs the LLM to respond with "none" for OOD nodes.
\vspace{-0.7cm}
\subsubsection{Cost}
\begin{table}[H]
\caption{The cost (dollars) of different LLMs for annotating 200 nodes on Cora dataset.}
\label{Table:cost}
\centering
\begin{adjustbox}{max width=\linewidth}
\begin{tabular}{lcccccc}
\hline
             & \textbf{GPT-3.5-turbo} & \textbf{GPT-4o-mini} & \textbf{GPT-4o} & \textbf{GPT-4} & \textbf{ds-v3} & \textbf{ds-r1}\\ \hline
\textbf{Cost}         &   0.07    &   0.02          &  0.50 & 3.70   & 0.03& 0.55     \\ \hline
\end{tabular}
\end{adjustbox}
\end{table}
\vspace{-0.5cm}
We randomly select 200 nodes from the Cora dataset and have different LLMs annotate them. The costs associated with each LLM are shown in Table \ref{Table:cost}. As observed, GPT-4o-mini incurs the lowest cost while achieving significantly better zero-shot open-world annotation performance than GPT-3.5-turbo. Therefore, in this paper, we use GPT-4o-mini for node annotation to reduce human costs in open-set scenarios.

\section{Conclusion and Future Directions}
In this paper, we introduce a novel approach leveraging the powerful zero-shot learning capabilities of LLMs for label-efficient graph OOD detection. We propose a general framework, \method, which filters out a large number of OOD nodes before annotation, significantly reducing human labeling costs. Additionally, \method utilizes zero-shot annotations from LLMs to train a lightweight GNN filter, further minimizing the reliance on LLMs. Unlike traditional multi-round active learning methods, our approach achieves high performance with a single round of node selection. A potential future research direction is to investigate more effective ways to leverage both clean and noisy labels to train a more noise-resistant ID classifier for graph OOD detection. Additionally, it would be interesting to explore whether in-context learning can improve node-level OOD detection performance compared to zero-shot OOD detection with LLMs.

\bibliographystyle{splncs04}
\bibliography{mybibliography}

\newpage 
\section*{Appendices}
\subsection*{Appendix A}
\label{appendix:metrics}

\textbf{AUROC} stands for "Area Under the Receiver Operating Characteristic Curve." It is a performance metric used in binary classification tasks to evaluate how well a model distinguishes between positive and negative classes. A higher AUROC value indicates better model performance in distinguishing between the two classes.

\noindent\textbf{AUPR} stands for the area under the precision-recall (PR) curve, similar to AUC, but it provides a more effective performance evaluation for imbalanced data.

\noindent\textbf{FPR95} refers to the false positive rate (FPR) when the true positive rate (TPR) reaches 95\%. It is used to measure the likelihood that an OOD example is incorrectly classified as ID when most ID samples are correctly identified. A lower FPR95 value indicates better detection performance.

\subsection*{Appendix B}
\label{appen:OOD Split}

\begin{table}[ht]
\caption{ID classes for different datasets.}
\label{table1}
\centering
\setlength{\tabcolsep}{1.9pt} 

\begin{tabular}{lc}
\hline
\textbf{Dataset}       & \textbf{ID classes}           \\ \hline
Cora                   & [2, 4, 5, 6]                       \\ 
Citeseer        & [0, 1, 2]                          \\ 
WikiCS           & [1, 4, 5, 6]                       \\ 
Pubmed             & [0, 1]               \\ \hline
\end{tabular}
\end{table}

\subsection*{Appendix C}
\begin{table}[H]
\caption{Proportion of annotated ID nodes under the label budget of $10\times K$.}
\label{Table:proportion}
\centering
\begin{adjustbox}{max width=\linewidth}
\begin{tabular}{lcccc}
\hline
\textbf{}              & \textbf{Cora}  & \textbf{Citeseer} & \textbf{Pubmed} & \textbf{Wiki-CS} \\ \hline
\textbf{Random}     &    0.4650      &    0.5400     &   0.6700    &    0.3650      \\
\textbf{Uncertainty}  &     0.3000     &  0.4000       &   0.5800     &  0.3100      \\ 
\textbf{FeatProp} &    0.2950     &  0.4000            &   0.4900       &    0.3250           \\
\textbf{MITIGATE} &    0.3000     &  0.4667            &   0.5300       &    0.2800           \\
\textbf{\method} &    0.6700     &  \textbf{0.9000}            &   0.8700       &    \textbf{0.8850}           \\
\textbf{\method-f} &    \textbf{0.7600}     &  0.8800            &   \textbf{0.8900}       &    0.7550           \\
\hline
\end{tabular}
\end{adjustbox}
\end{table}

\newpage
\subsection*{Appendix D}
The prompt designed for zero-shot OOD detection is shown in the following Box.

\begin{tcolorbox}[width=\columnwidth, colback=green!10, colframe=gray, title=Zero-Shot OOD Detection and ID Classification Prompt, fonttitle=\bfseries]

\label{box:annotation_long_prompt}

You are an expert text classification assistant specializing in identifying whether a given \textbf{\{object\}} belongs to the predefined in-distribution categories or is out-of-distribution (OOD).

A \textbf{\{object\}} is considered as \textbf{out-of-distribution (OOD)} if it does NOT belong to \textbf{any of the in-distribution category(ies)} listed below.

Your task is, given the content of the \textbf{\{object\}} below, to determine whether it is an out-of-distribution (OOD) \textbf{\{object\}}. If it is an OOD \textbf{\{object\}}, answer \texttt{"none"}. If it is not an OOD \textbf{\{object\}}, determine which in-distribution category below it belongs to. Provide a brief explanation of your reasoning and assign a confidence score between 0 and 1 for your justification.

\textbf{In-distribution Categories:}
\vspace{-0.4cm}

\begin{itemize}
    \item \textbf{Category 1}
    \item \textbf{Category 2}
    \item \textbf{...} 
\end{itemize}
\vspace{-0.4cm}

If you are uncertain whether the \textbf{\{object\}} significantly aligns with any of the in-distribution category(ies), assume that it does NOT align with them, which means it is an out-of-distribution \textbf{\{object\}}.

\textbf{The description of the \textbf{\{object\}} that you need to identify is as follows:}
\texttt{[Insert \textbf{\{Object\}} Content Here]}
\end{tcolorbox}
%





\end{document}